\documentclass[]{bytedance_seed}

\usepackage[toc,page,header]{appendix}

\usepackage{natbib}

\usepackage{CJKutf8}

\usepackage{xargs}  

\usepackage{todonotes}  

\usepackage{multirow}

\usepackage{cleveref}

\usepackage{amsmath}
\usepackage{dsfont}

\usepackage{subcaption}

\usepackage{svg}

\usepackage{mathrsfs}
\usepackage{adjustbox}
\usepackage{multirow}
\usepackage{multirow}
\usepackage{multicol}
\usepackage{tcolorbox}
\usepackage{changepage}
\usepackage{graphicx}
\usepackage{amssymb}
\usepackage{array}
\usepackage{bm}
\usepackage{hyperref}
\usepackage{minitoc}

\title{Seedance 1.5 pro: A Native Audio-Visual Joint Generation Foundation Model}

\author[]{ByteDance Seed}
\abstract{
Recent strides in video generation have paved the way for unified audio-visual generation. In this work, we present Seedance 1.5 pro, a foundational model engineered specifically for native, joint audio-video generation. 
Leveraging a dual-branch Diffusion Transformer architecture, the model integrates a cross-modal joint module with a specialized multi-stage data pipeline, achieving exceptional audio-visual synchronization and superior generation quality. To ensure practical utility, we implement meticulous post-training optimizations, including Supervised Fine-Tuning (SFT) on high-quality datasets and Reinforcement Learning from Human Feedback (RLHF) with multi-dimensional reward models. Furthermore, we introduce an acceleration framework that boosts inference speed by over 10$\times$.
Seedance 1.5 pro distinguishes itself through precise multilingual and dialect lip-syncing, dynamic cinematic camera control, and enhanced narrative coherence, positioning it as a robust engine for professional-grade content creation.
Seedance 1.5 pro is now accessible on \href{https://console.volcengine.com/ark/region:ark+cn-beijing/experience/vision?type=GenVideo}{Volcano Engine}\textsuperscript{$\alpha$}.
}

\checkdata[Official Page]{\url{https://seed.bytedance.com/seedance1_5_pro} }
\checkdata[\textsuperscript{$\alpha$}Model ID]{Doubao-Seedance-1.5-pro}

\begin{document}
\begin{CJK*}{UTF8}{gbsn}

\maketitle

\definecolor{chinese_red}{HTML}{8B4513}
\definecolor{english_blue}{HTML}{4169E1}

\begin{figure}[ph]
\begin{center}

\includegraphics[height=8.0cm]{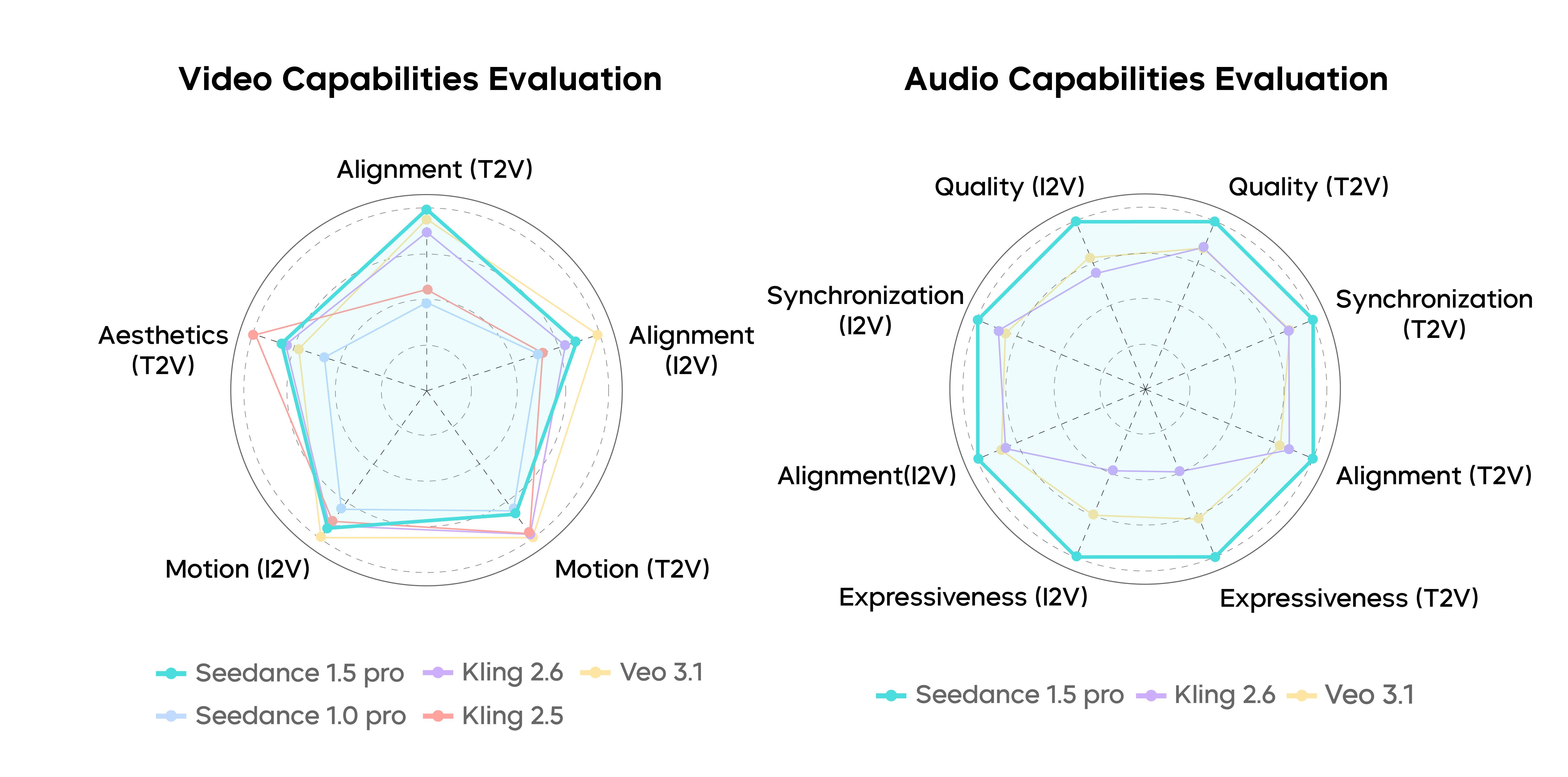}
\end{center}
\label{fig:overall_eval}
\caption{Overall Evaluation. Left: Video Evaluation; Right: Audio Evaluation.}
\end{figure}%
% \footnotetext[1]{
% Seedream 3.0 ranks first at Artifcial Analysis Text to Image Model Leaderboard with an Arena ELO score of 1158 at 17.0K Appearances. For the latest information, please visit
% \url{https://artificialanalysis.ai/text-to-image/arena?tab=Leaderboard}}

%\begin{figure}[pt]
%\begin{center}
%\includegraphics[width=0.88\linewidth]{figures/seedream3_overall.png}
%\end{center}
%\label{fig:teaser}
%\vspace{-1pt}
%\caption{Seedream 3.0 visualization.}
%\end{figure}

% \newpage

\clearpage

% \tableofcontents

% \newpage

\section{Introduction}
Over the past year, the field of visual (video) generation  \cite{gong2025seedream, gao2025seedream, seedream2025seedream,wu2025qwen,wu2025hunyuanvideo,wan2025wan}
 has seen rapid development. The emergence of proprietary commercial systems, such as Veo, Sora, Kling series and Seedance \cite{gao2025seedance}, alongside open-source models like Wan~\citep{wan2025wan} and Hunyuan Video 1.5~\citep{kong2024hunyuanvideo}, has significantly propelled the widespread adoption of video generation technologies across both academia and industry. More recently, substantial progress has been made in joint audio-video generation. The release of Wan 2.5, Kling 2.6, and Sora 2 marks a solid step forward, transforming video generation capabilities into practical, utility-driven tools.

In this work, we present Seedance 1.5 pro, a foundational model designed with native support for joint video-audio generation. It is capable of performing a diverse range of tasks, including text-to-audio-video synthesis and image-guided audio-video generation. Seedance 1.5 pro incorporates the following key technical advancements:
\begin{itemize}[leftmargin=*]

\item \textbf{\textit{Comprehensive Audio–Visual Data Framework.}}
We present a holistic data framework for high-quality video–audio generation that integrates a multi-stage curation pipeline, an advanced captioning system, and scalable infrastructure. The pipeline prioritizes video–audio coherence, motion expressiveness, and curriculum-based data scheduling, while our captioning system provides rich, professional-grade descriptions for both video and audio modalities. This robust framework is underpinned by an efficient engineering infrastructure optimized for massive multi-modal data processing.

\item \textbf{\textit{Unified Multimodal Joint Generation Architecture.}}
To achieve native video–audio joint synthesis, we propose a unified framework based on the MMDiT \cite{esser2024scaling} architecture. This design facilitates deep cross-modal interaction, ensuring precise temporal synchronization and semantic consistency between visual and auditory streams. By leveraging multi-task pre-training on large-scale mixed-modality datasets, our model achieves robust generalization across diverse downstream tasks, including Text-to-Video-Audio (T2VA), Image-to-Video-Audio (I2VA), and unimodal video generation (T2V, I2V).

 \item \textbf{\textit{Meticulous Post-training Optimization.}}  
We utilized high-quality audio-video datasets for Supervised Fine-Tuning (SFT), followed by a Reinforcement Learning from Human Feedback (RLHF\cite{liu2025flow,xue2025dancegrpo,wu2025rewarddance,zhang2024unifl}) algorithm specifically tailored for audio-video contexts. Specifically, our multi-dimensional reward model enhances performance in Text-to-Video (T2V) and Image-to-Video (I2V) tasks, improving motion quality, visual aesthetics, and audio fidelity. Moreover, targeted infrastructure optimizations to our RLHF pipeline have yielded a nearly 3$\times$ improvement in training speed.

\item \textbf{\textit{Efficient Inference Acceleration.}} We further optimized a multi-stage distillation framework \cite{ren2025hyper,geng2025mean,shao2025rayflow}  to substantially reduce the Number of Function Evaluations (NFE) required during generation. By integrating inference infrastructure optimizations—such as quantization and parallelism, we achieved an end-to-end acceleration exceeding 10× while preserving model performance.

\end{itemize} 

\begin{figure*}[hb]
\centering
\includegraphics[width=\textwidth]{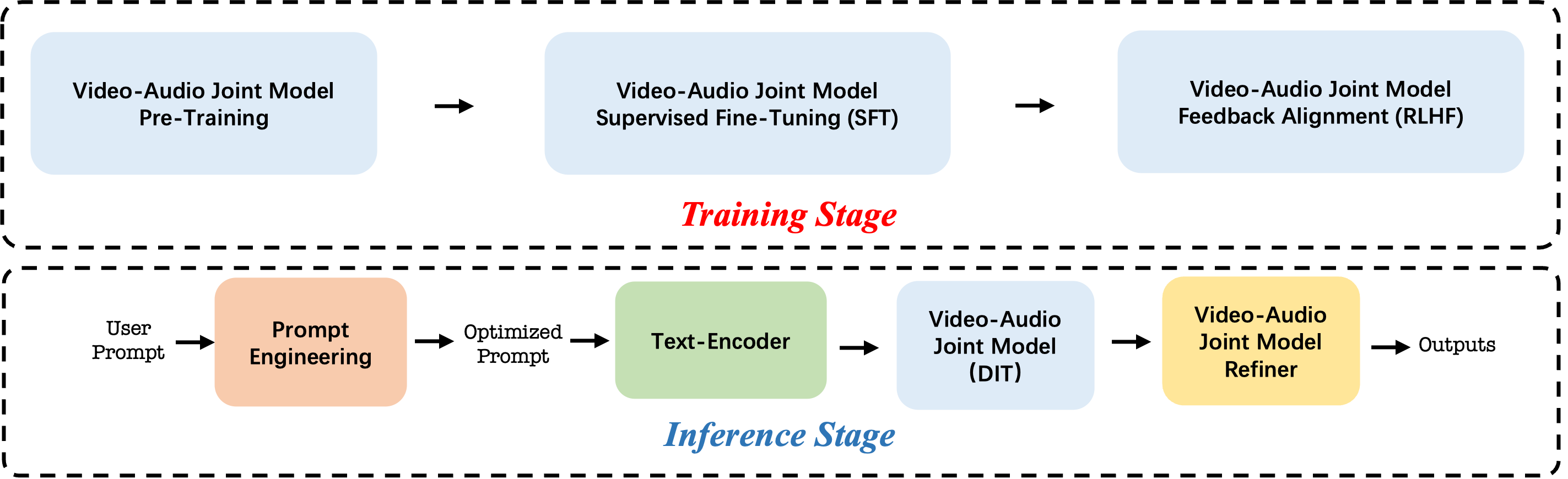}
\caption{Overview of training and inference pipeline of Seedance 1.5 pro.}
\label{fig:stage}
\end{figure*}

Seedance 1.5 pro aims to elevate the ceiling of visual impact and motion expressiveness. Seedance 1.5 pro demonstrates comprehensive advancements in multilingual adaptation, motion expressiveness, camera scheduling, and dialect delivery. Its performance in scenarios such as Chinese film production, short dramas, and traditional performing arts demonstrates significant potential for practical application in multi-shot video generation workflows:

\begin{itemize}[leftmargin=*]

\item \textbf{Precise Audio-Visual Synchronization and Multilingual Support.}
The model realizes high audio-visual consistency during generation, significantly improving the alignment accuracy of lip movements, intonation, and performance rhythm. It natively supports multiple languages and regional dialects, accurately capturing their unique vocal prosody and emotional tension. This capability imparts a natural performance quality to stylized narratives, such as comedies and animations, thereby greatly enhancing character atmosphere and expressiveness.

\item \textbf{Cinematic Camera Control and Dynamic Tension.} 
The model possesses autonomous camera scheduling capabilities, enabling the execution of complex movements such as continuous long takes and dolly zooms (Hitchcock zoom). Additionally, it achieves cinematic scene transitions and professional color grading, substantially enhancing the dynamic tension of the video.

\item \textbf{Enhanced Semantic Understanding and Narrative Coherence.} 
Through strengthened semantic understanding, Seedance 1.5 pro achieves precise analysis of narrative contexts. It significantly improves the overall narrative coordination of audio-visual segments, providing strong support for professional-grade content creation.

\end{itemize}

Seedance 1.5 pro is scheduled to be integrated into multiple platforms, including Doubao\footnotemark[1] and Jimeng\footnotemark[2], by December 2025. We envision this model serving as a vital productivity tool, designed to enhance both professional workflows and daily creative applications.

\footnotetext[1]{https://www.doubao.com/chat/create-video}
\footnotetext[2]{https://jimeng.jianying.com/ai-tool/video/generate}

% \newpage

% \input{sections/Model_Design}

% \input{sections/Data}

% \input{sections/Model_Train}

% \input{sections/Model_Optimization}

% \input{sections/infrastructure}

\section{Evaluation}
As AIGC video generation evolves toward deeper multimodal integration, the incorporation of audio capabilities has become a pivotal factor in transforming video generation from high-quality visual assets into holistic and production-ready works. By achieving precise audio–visual synchronization and coherent emotional alignment, the model outputs are no longer fragmented visuals, but represent increasingly mature productions characterized by narrative completeness and immersive quality. 

This chapter provides a comprehensive analysis of the Seedance 1.5 pro model. Section 2.1 details the internal evaluation, outlining our assessment methodology in both video and audio dimensions and presenting comparative analyses against state-of-the-art (SOTA) video generation models. Finally, we highlight the distinct advantages of the model in diverse application scenarios.
 
\subsection{Comprehensive Evaluation}
In contrast to third-party benchmarks that prioritize general user preference, we established a comprehensive, multi-dimensional framework designed to assess model performance across diverse domains and capabilities. By integrating rigorous benchmark construction with specific evaluation criteria, this framework provides granular insights that guide model development and facilitate rapid iteration.
Building upon SeedVideoBench-1.0, which was grounded in the analysis of real-world user prompts, we introduce the enhanced SeedVideoBench-1.5. Relative to its predecessor, this version significantly expands the coverage of industry-specific scenarios, such as advertising and micro-dramas, while incorporating advanced metrics for audio evaluation. Consistent with our established methodology, we collaborated with professional film directors to codify these criteria and engaged experts from film production, cinematography, and design to conduct expert-level human assessments.

% \begin{figure*}[t]
% \centering
% \includegraphics[width=\linewidth]{figures/t2v_all_per_dim.pdf}
% \caption{\textbf{placeholder} Absolute Evaluation for Text-to-Video task.}
% \label{fig:t2v_abs}
% \end{figure*}

% \begin{figure*}[!h]
% \centering
% \includegraphics[width=\linewidth]{figures/t2v_gsb_test.pdf}
% \caption{\textbf{placeholder} GSB Evaluation for Text-to-Video Task.}
% \label{fig:t2v_gsb}
% \end{figure*}

\subsubsection{SeedVideoBench 1.5}
For the video dimension, we provide a detailed taxonomy of evaluation cases and attribute labels covering core aspects such as subjects, motion dynamics, interactions, and camera movements. The updated benchmark also integrates labels for diversified application scenarios, including advertising, social media content, and short-form narrative content.
In light of the Seedance 1.5 capability for joint audio–video generation, we have upgraded the evaluation framework to include a comprehensive audio dimension. The primary label categories are defined as follows:
\begin{itemize}[leftmargin=*]
\item ~\underline{\textit{Human Voice Types:}}  As a critical component of video creation, this category encompasses speech, singing, and non-verbal vocalizations (e.g., laughter). SeedVideoBench 1.5 introduces a comprehensive taxonomy with fine-grained sub-dimensions to capture the diverse requirements of human voice generation.
\item ~\underline{\textit{Human Voice Attributes:}}   These labels characterize specific vocal qualities, including timbre, accent, and emotional tone.
\item ~\underline{\textit{Non-Speech Audio (Sound Effects/Music): }} This category includes all environmental and musical elements essential for perceptual realism and scene coherence. The labeling system classifies audio by source (e.g., animals, mechanical tools), acoustic properties, musical genres, and technical parameters.
\end{itemize}

The audio labeling methodology follows consistent principles across both T2V and I2V  tasks. However, in I2V contexts, audio synthesis is explicitly conditioned on the visual cues present in the reference image to ensure semantic consistency and cross-modal coherence.

\subsubsection{Video Evaluation Metrics}
In SeedVideoBench 1.5, we extend our prior focus on motion dynamics, prompt adherence, aesthetic quality, and subject consistency by introducing specialized evaluation metrics designed to better reflect professional production requirements. The key updates are detailed below.

\begin{itemize}[leftmargin=*]
\item ~\underline{\textit{Motion Quality:}} Motion quality remains the most immediate and critical concern for users. While baseline attributes—such as stability, physical plausibility, and temporal accuracy—remain mandatory, this benchmark iteration places renewed emphasis on Video Vividness, which has emerged as an increasingly important indicator for downstream professional applications. As generative capabilities mature across the field, foundational motion metrics are likely to approach performance saturation. Consequently, practitioners in advertising and film production increasingly demand video outputs that exhibit higher levels of vividness across diverse scenarios.
Notably, among several state-of-the-art models, we observe a common trade-off in which slow-motion generation is employed to artificially enhance perceived stability, a strategy that significantly degrades motion vividness and expressive quality.
To address this limitation, we evaluate video vividness as a composite perceptual metric across four principal dimensions: action, camera movement, atmosphere, and emotion. The following two were in particular examined:

\noindent \textbf{Action Dimension:} Vividness in action is characterized by nuanced facial expressions, aesthetically refined body poses, high-fidelity reproduction of fine-grained motions, and realistic interactions with the environment. Together, these factors determine the perceptual authenticity and emotional expressiveness of generated motion.

\noindent\textbf{Camera Dimension:} Cinematic composition and dynamic camera movements play a critical role in enhancing visual expressiveness, strengthening emotional delivery, and supporting temporal narrative coherence across shots.

\item ~\underline{\textit{Prompt Following:}} While strict adherence to user instructions remains necessary when users explicitly request word-for-word execution, the definition of prompt adherence has been updated to reflect the improved semantic understanding and intent recognition of modern generative models. Rather than emphasizing surface-level keyword matching, the evaluation now prioritizes consistency with the user’s underlying intent.

In narrative-driven social media scenarios, models are permitted a degree of intent-aligned creative flexibility, provided that the core user intention is preserved. Such flexibility may include completing missing visual details, refining narrative structure, or generating dialogue that better aligns with the intended emotional tone—thereby contributing to stronger storytelling quality and cross-modal expressive coherence.

\subsubsection{Audio Evaluation Metrics}

Our audio evaluation metrics are designed to quantitatively characterize model performance across multiple dimensions, establishing a roadmap for advancing audio generation toward cinematic-level production standards. The evaluation framework comprises four core components:

\item ~\underline{\textit{Audio Prompt Following:}} Analogous to text-video alignment, this metric assesses the fidelity with which vocal elements, dialogue, and sound effects adhere to user instructions and intended semantics, without semantic drift—even during creative extrapolation (i.e., intent-consistent content expansion). Common failure modes include the omission of specified sound effects, linguistic or dialectal inaccuracies, and audio–vision mismatches (e.g., speech generation without corresponding lip motion).

\item ~\underline{\textit{Audio Quality:}} This metric quantifies the intrinsic acoustic quality of the output, encompassing both vocal and non-vocal components. Key evaluation criteria include the presence of artifacts (e.g., clipping, truncation), spatial soundstage rendering, timbre realism, and overall signal clarity.

\item ~\underline{\textit{Audio–Visual Synchronization:}} This metric measures the temporal alignment between auditory and visual streams. It evaluates the synchronization of speech-to-lip dynamics (e.g., mitigating perceptual ventriloquism effects), the alignment of sound effects with visual events, and the presence of auditory cues corresponding to salient on-screen actions.

\item ~\underline{\textit{Audio Expressiveness:}}
This metric evaluates the extent to which audio augments the emotional resonance of the video. It considers the thematic appropriateness of background music (BGM), the emotional inflection of speech, and the audio’s contribution to atmospheric immersion and narrative coherence and depth.

\end{itemize}

% \begin{figure*}[t]
% \centering
% \includegraphics[width=\linewidth]{figures/i2v_all_per_dim.pdf}
% \vspace{-1cm}
% \caption{\textbf{placeholder} Absolute Evaluation for Image-to-Video task.}
% \label{fig:i2v_abs}
% \end{figure*}

% \begin{figure*}[!h]
% \centering
% \includegraphics[width=\linewidth]{figures/i2v_gsb_test.pdf}
% \caption{\textbf{placeholder} GSB Evaluation for Image-to-Video task.}
% \label{fig:i2v_gsb}
% \end{figure*}

\subsubsection{Human Evaluation}
\textbf{Video.} Leveraging the SeedVideoBench 1.5 framework, we conducted a comprehensive comparative evaluation of Seedance 1.5 pro against several state-of-the-art video generation models across both Text-to-Video (T2V) and Image-to-Video (I2V) tasks. The comparative baselines include Kling 2.5, Kling 2.6, Veo 3.1, and Seedance 1.0 Pro.
To ensure a robust assessment, we employed a dual-metric evaluation protocol:
\begin{itemize}
\item \textbf{Absolute Score:} This metric utilizes a 5-point Likert scale (ranging from 1 for ``Extremely Dissatisfied'' to 5 for ``Extremely Satisfied'') to facilitate a standardized performance comparison across different models.
\item \textbf{Good-Same-Bad (GSB):} This metric involves pairwise comparisons to assess relative video quality, thereby enabling a more granular differentiation of model outputs.
\end{itemize}

\begin{figure*}[h]
\centering
\includegraphics[width=\linewidth]{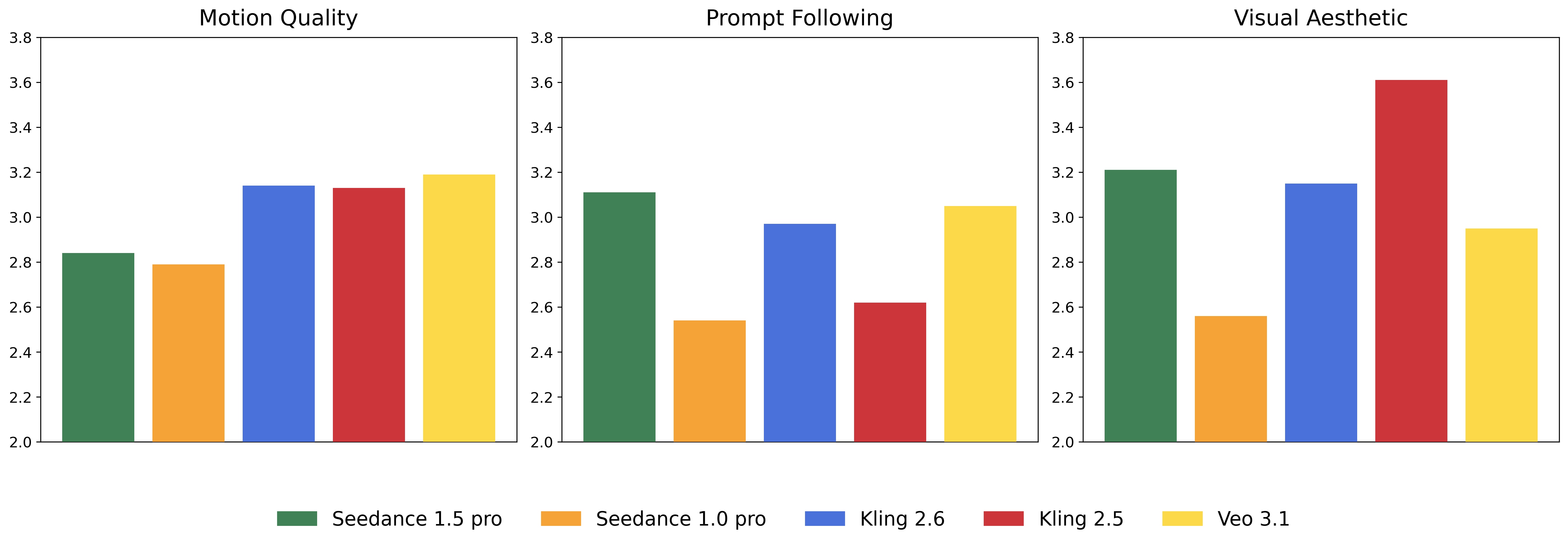}
\caption{Video Absolute Evaluation for Text-to-Video task.}
\label{fig:t2v_abs}
\end{figure*}

\begin{figure*}[h]
\centering
\includegraphics[width=\linewidth]{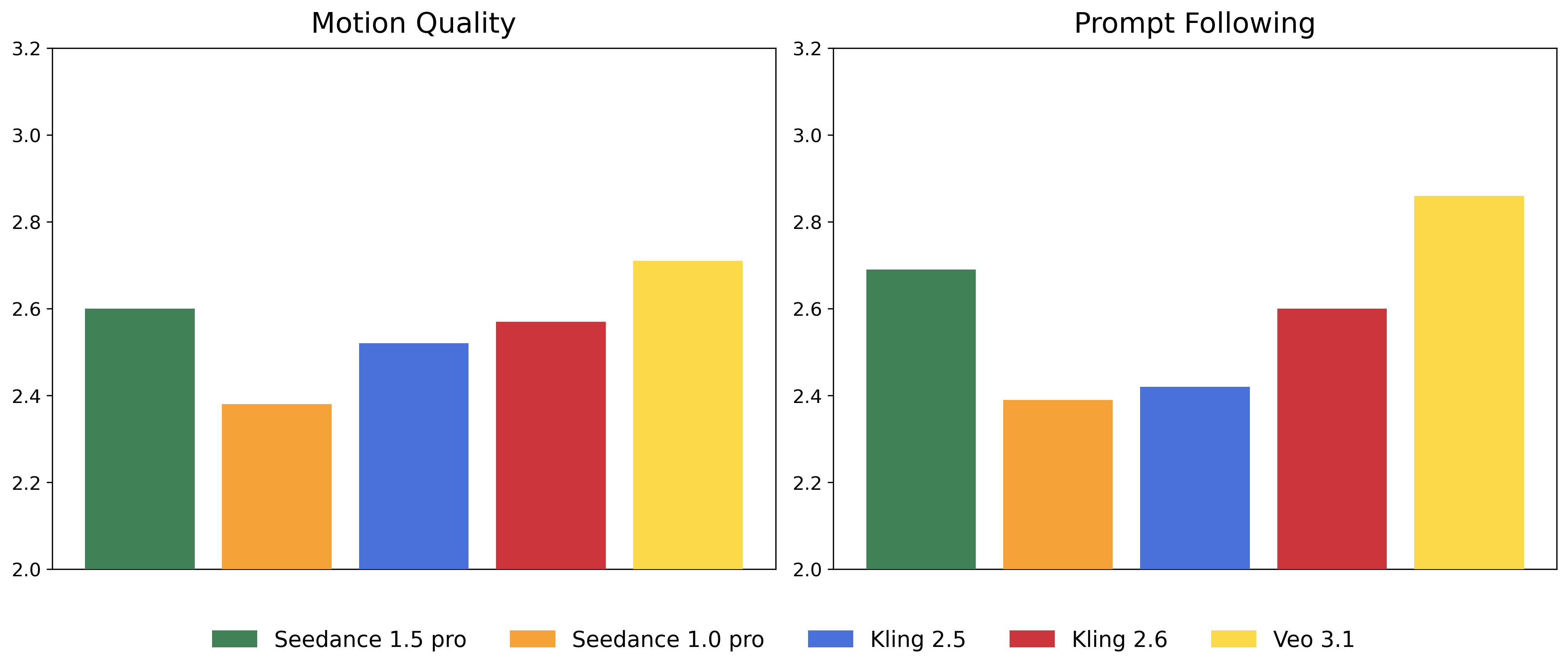}
\caption{Video Absolute Evaluation for Image-to-Video task.}
\label{fig:i2v_abs}
\end{figure*}

Figures \ref{fig:t2v_abs} and \ref{fig:i2v_abs} present the absolute evaluation scores of the video generation models in the Text-to-Video (T2V) and Image-to-Video (I2V) task. Seedance 1.5 pro demonstrates a significant improvement over its predecessor, Seedance 1.0 Pro. Specifically, in T2V generation, Seedance 1.5 pro achieves a leading position in instruction following (alignment). Furthermore, it exhibits strong competitiveness in terms of visual aesthetics and motion dynamics, as well as in Image-to-Video (I2V) tasks.

\begin{figure*}[h]
\centering
\includegraphics[width=\linewidth]{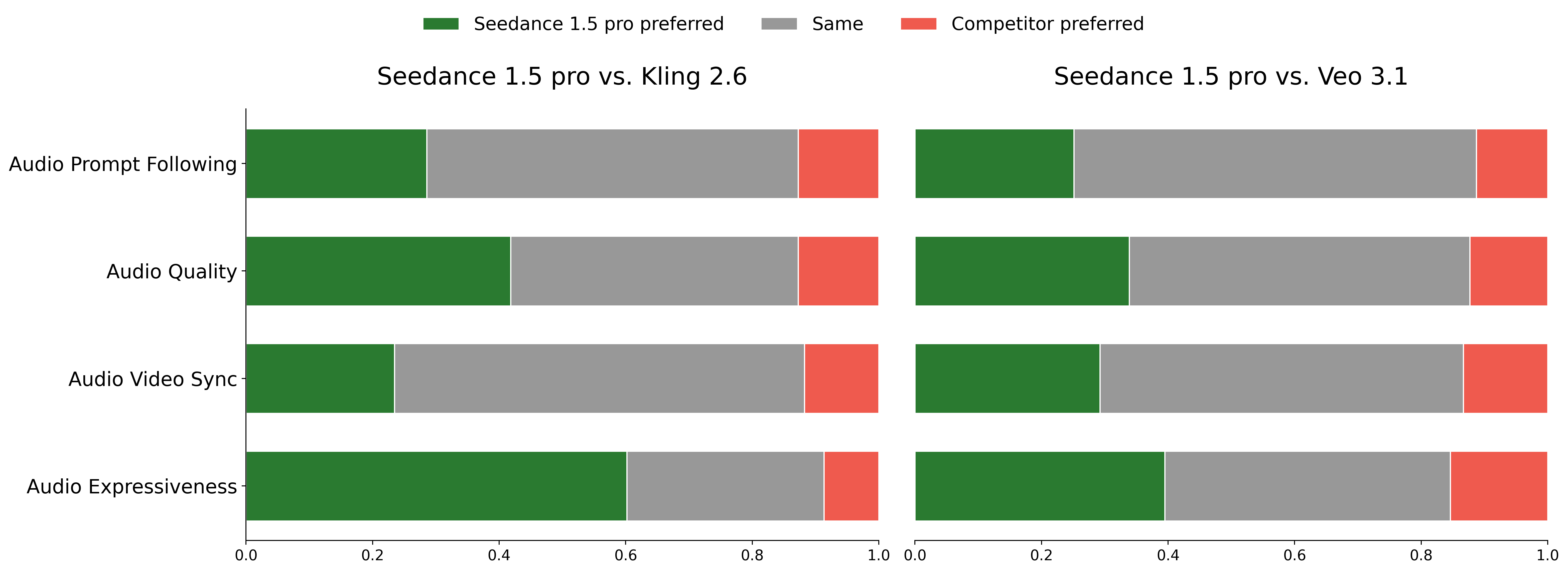}
\caption{Audio GSB Evaluation for Image-to-Video task.}
\label{fig:i2v_gsb}
\end{figure*}

\textbf{Audio.} Figure \ref{fig:t2v_gsb} and \ref{fig:i2v_gsb} presents a multi-dimensional side-by-side (GSB) comparative evaluation of audio performance between the Seedance 1.5 pro and several competing systems. While models such as Veo 3.1, Wan 2.5, Kling 2.6, and Sora 2 demonstrate robust audio-generation capabilities, the Seedance 1.5 pro exhibits distinct advantages in several key dimensions:

\begin{figure*}[h]
\centering
\includegraphics[width=\linewidth]{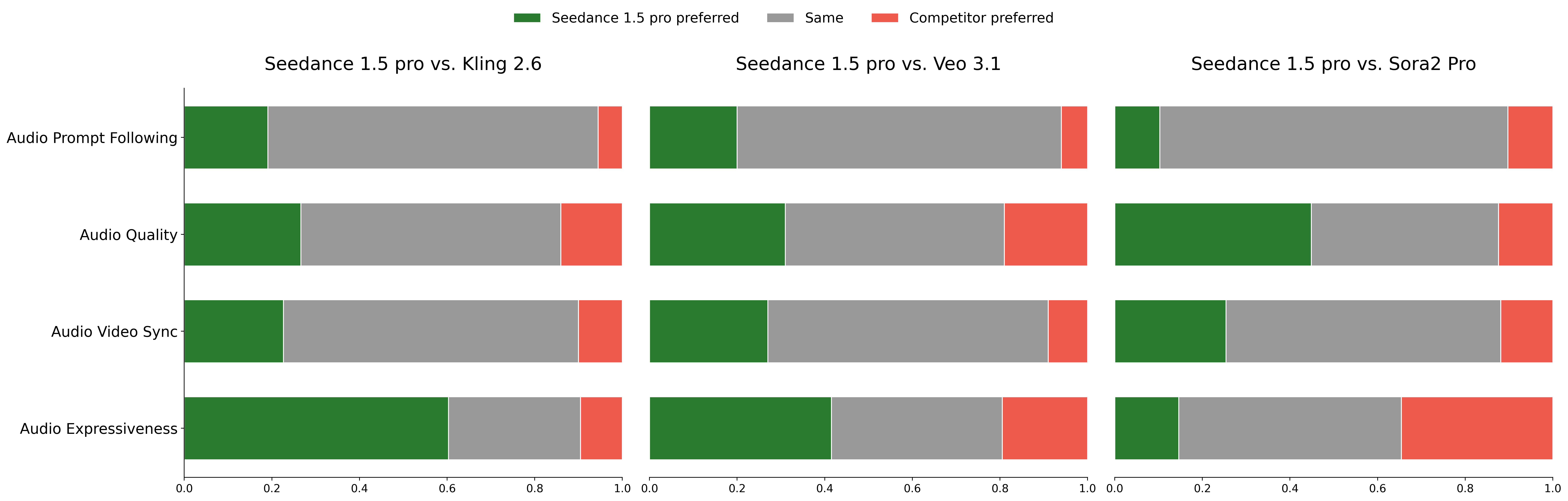}
\caption{Audio GSB Evaluation for Text-to-Video task.}
\label{fig:t2v_gsb}
\end{figure*}

\noindent\textbf{Superiority in Chinese-Language Audio.} Seedance 1.5 pro consistently outperforms Veo 3.1 in Chinese vocal generation. It demonstrates a clear advantage in synthesizing dialogue, dialects, and monologues within Chinese-language contexts, reliably producing accurate responses with high articulatory clarity. Notably, the model remains largely free from common artifacts such as syllable dropping or mispronunciation.

\noindent\textbf{Enhanced Audio–Visual Synchronization.} Seedance 1.5 pro model excels in aligning both vocal tracks and sound effects with visual cues. In terms of lip–audio synchronization, the model accurately corresponds to the number and identity of speaking characters, effectively mitigating errors related to mouth motion redundancy or omission that typically result in audio–visual temporal misalignment. In this regard, its performance surpasses both Veo 3.1 and Kling 2.6.

\noindent\textbf{Comparative Analysis of Audio Expressiveness.}  Sora 2 demonstrates strong competence in emotional expressiveness, delivering particularly vivid emotional inflection in its audio outputs. By contrast, Seedance 1.5 pro maintains a more balanced and controlled expressiveness profile, achieving consistent emotional alignment with visual content while avoiding over-exaggeration. This characteristic is especially advantageous in professional production scenarios that require stable tone control and narrative coherence.

\subsection{Potential Advantages in Application Scenarios}

Seedance 1.5 pro further strengthens its capabilities in character expressiveness and stylized visual construction, with particularly strong performance in Chinese-language contexts. Compared to previous generations, Seedance 1.5 pro demonstrates substantial improvements in character dialogue, including support for multiple Chinese dialects, as well as in the execution of complex camera dynamics. These advancements make the model well-suited for Chinese film production, short-form micro-dramas, and theatrical storytelling scenarios.

Specifically, Seedance 1.5 pro maintains consistent lip synchronization, vocal tonality, and performance rhythm across continuous shots. The model exhibits robust performance in dialect-rich settings, such as Sichuanese, Taiwan Mandarin, Cantonese, and Shanghainese, producing natural prosody and speech patterns that closely resemble authentic regional usage. These capabilities are particularly beneficial for genre-specific storytelling, including comedy and comic-style narratives, where dialectical cadence and performative tension play a critical role in establishing atmosphere and comedic timing.

In terms of visual composition, Seedance 1.5 pro demonstrates mature control over complex camera operations. The model reliably executes orbital, arc, and tracking shots while preserving visual style consistency between generated sequences and reference imagery, thereby ensuring cinematic visual continuity across scenes. Moreover, leveraging improved prompt understanding, Seedance 1.5 pro can autonomously introduce novel subjects and actions that remain aligned with the narrative genre, enhancing overall scene coherence and immersive continuity.

\begin{figure*}[h]
\centering
\includegraphics[width=\linewidth]{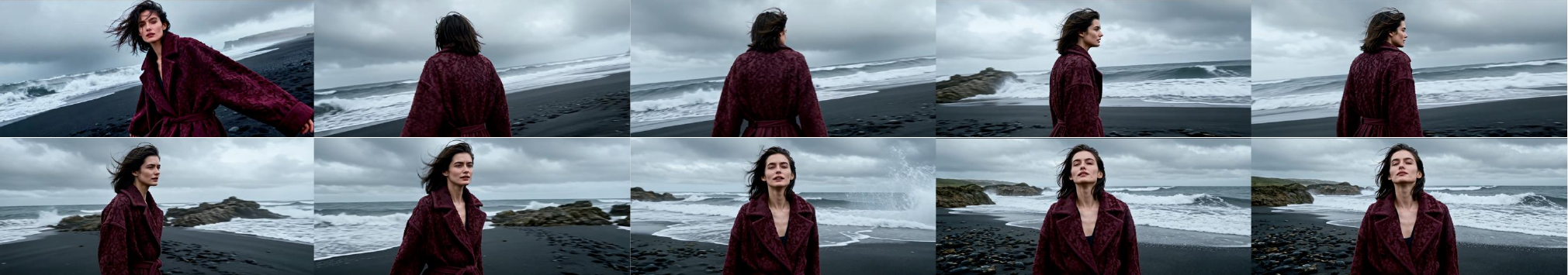}
\label{fig:multi_shot_woman}
\end{figure*}

Additionally, Seedance 1.5 pro demonstrates a heightened sensitivity to the traditional Chinese performance system within stage opera scenarios. While its mastery of specific vocal styles across different opera sub-genres is still evolving, the model is already capable of capturing the distinctive cadence and flavor of operatic speech (Nianbai). By integrating nuanced performance details, such as the orchid hand gesture (Lanhua Zhi) and the stylized eye expressions typical of comedic roles, Seedance 1.5 pro effectively constructs a performance atmosphere deeply rooted in Eastern operatic aesthetics.

\begin{figure*}[h]
\centering
\includegraphics[width=\linewidth]{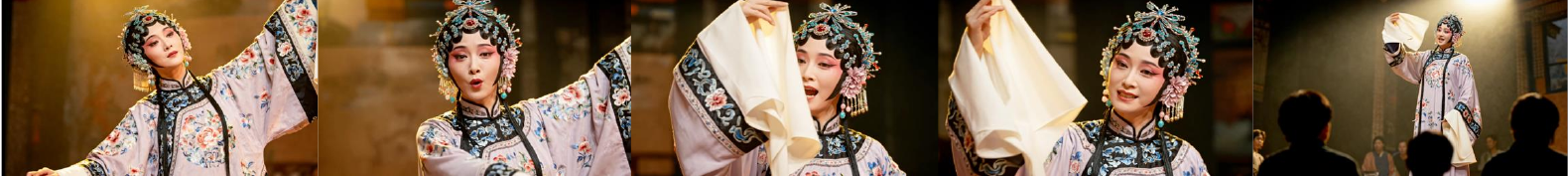}
\label{fig:multi_shot_woman}
\end{figure*}
In realistic cinematic close-up shots, Seedance 1.5 pro sustains emotional continuity through subtle and coherent facial micro-expressions. Even in segments with minimal dialogue, the model preserves the integrity of character performance, providing creators with greater narrative flexibility and space for interpretive silence.

\begin{figure*}[h]
\centering
\includegraphics[width=\linewidth]{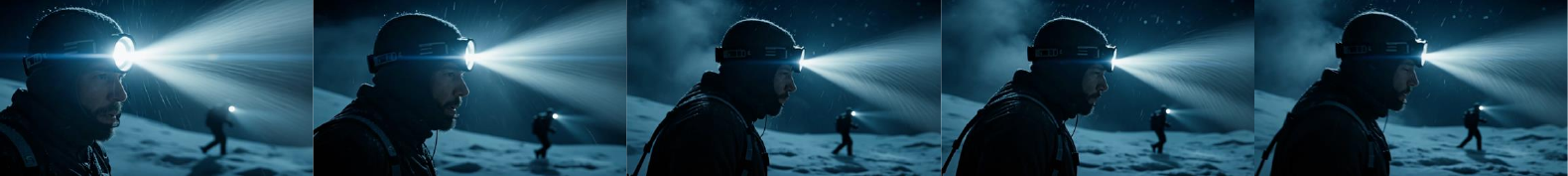}
\label{fig:multi_shot_woman}
\end{figure*}

\begin{figure*}[h]
\centering
\includegraphics[width=\linewidth]{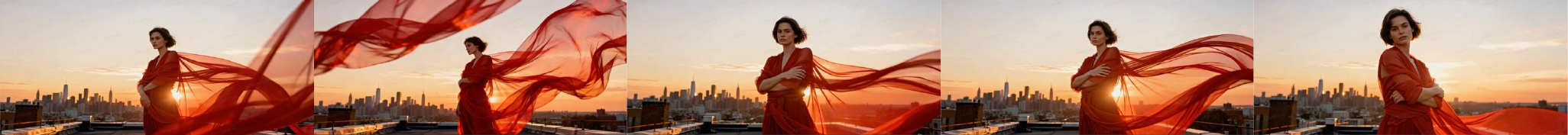}
\label{fig:multi_shot_woman}
\end{figure*}

In summary, the advance of Seedance 1.5 pro in Chinese-language generation, stylized visual rendering, complex camera control, and traditional theatrical expression enable a high degree of narrative expressiveness and seamless audio–visual integration. Collectively, these capabilities enhance the creative controllability of the model in various application domains, including Chinese film production, short-form drama creation, and opera-inspired audiovisual storytelling.

\clearpage

\bibliographystyle{plainnat}
\bibliography{main}

\clearpage

\beginappendix
\section{Contributions and Acknowledgments}
\label{contributions}

All Authors of Seedance are listed in alphabetical order by their last names.

\begin{multicols}{3} %
\sffamily{\color{seedblue}  \large{Authors}} \\
\\
\color{seedblue}Heyi Chen\\
\color{seedblue}Siyan Chen\\
\color{seedblue}Xin Chen\\
\color{seedblue}Yanfei Chen\\
\color{seedblue}Ying Chen\\
\color{seedblue}Zhuo Chen\\
\color{seedblue}Feng Cheng\\
\color{seedblue}Tianheng Cheng\\
\color{seedblue}Xinqi Cheng\\
\color{seedblue}Xuyan Chi\\
\color{seedblue}Jian Cong\\
\color{seedblue}Jing Cui\\
\color{seedblue}Qinpeng Cui\\
\color{seedblue}Qide Dong\\
\color{seedblue}Junliang Fan\\
\color{seedblue}Jing Fang\\
\color{seedblue}Zetao Fang\\
\color{seedblue}Chengjian Feng\\
\color{seedblue}Han Feng\\
\color{seedblue}Mingyuan Gao\\
\color{seedblue}Yu Gao\\
\color{seedblue}Dong Guo\\
\color{seedblue}Qiushan Guo\\
\color{seedblue}Boyang Hao\\
\color{seedblue}Hongxiang Hao\\
\color{seedblue}Qingkai Hao\\
\color{seedblue}Bibo He\\
\color{seedblue}Qian He\\
\color{seedblue}Tuyen Hoang\\
\color{seedblue}Ruoqing Hu\\
\color{seedblue}Xi Hu\\
\color{seedblue}Weilin Huang\\
\color{seedblue}Zhaoyang Huang\\
\color{seedblue}Zhongyi Huang\\
\color{seedblue}Donglei Ji\\
\color{seedblue}Jianwen Jiang\\
\color{seedblue}Siqi Jiang\\
\color{seedblue}Yunpu Jiang\\
\color{seedblue}Zhuo Jiang\\
\color{seedblue}Ashley Kim\\
\color{seedblue}Jianan Kong\\
\color{seedblue}Zhichao Lai\\
\color{seedblue}Shanshan Lao\\
\color{seedblue}Yichong Leng\\
\color{seedblue}Ai Li\\
\color{seedblue}Feiya Li\\
\color{seedblue}Gen Li\\
\color{seedblue}Huixia Li\\
\color{seedblue}JiaShi Li\\
\color{seedblue}Liang Li\\
\color{seedblue}Ming Li\\
\color{seedblue}Shanshan Li\\
\color{seedblue}Tao Li\\
\color{seedblue}Xian Li\\
\color{seedblue}Xiaojie Li\\
\color{seedblue}Xiaoyang Li\\
\color{seedblue}Xingxing Li\\
\color{seedblue}Yameng Li\\
\color{seedblue}Yifu Li\\
\color{seedblue}Yiying Li\\
\color{seedblue}Chao Liang\\
\color{seedblue}Han Liang\\
\color{seedblue}Jianzhong Liang\\
\color{seedblue}Ying Liang\\
\color{seedblue}Zhiqiang Liang\\
\color{seedblue}Wang Liao\\
\color{seedblue}Yalin Liao\\
\color{seedblue}Heng Lin\\
\color{seedblue}Kengyu Lin\\
\color{seedblue}Shanchuan Lin\\
\color{seedblue}Xi Lin\\
\color{seedblue}Zhijie Lin\\
\color{seedblue}Feng Ling\\
\color{seedblue}Fangfang Liu\\
\color{seedblue}Gaohong Liu\\
\color{seedblue}Jiawei Liu\\
\color{seedblue}Jie Liu\\
\color{seedblue}Jihao Liu\\
\color{seedblue}Shouda Liu\\
\color{seedblue}Shu Liu\\
\color{seedblue}Sichao Liu\\
\color{seedblue}Songwei Liu\\
\color{seedblue}Xin Liu\\
\color{seedblue}Xue Liu\\
\color{seedblue}Yibo Liu\\
\color{seedblue}Zikun Liu\\
\color{seedblue}Zuxi Liu\\
\color{seedblue}Junlin Lyu\\
\color{seedblue}Lecheng Lyu\\
\color{seedblue}Qian Lyu\\
\color{seedblue}Han Mu\\
\color{seedblue}Xiaonan Nie\\
\color{seedblue}Jingzhe Ning\\
\color{seedblue}Xitong Pan\\
\color{seedblue}Yanghua Peng\\
\color{seedblue}Lianke Qin\\
\color{seedblue}Xueqiong Qu\\
\color{seedblue}Yuxi Ren\\
\color{seedblue}Kai Shen\\
\color{seedblue}Guang Shi\\
\color{seedblue}Lei Shi\\
\color{seedblue}Yan Song\\
\color{seedblue}Yinglong Song\\
\color{seedblue}Fan Sun\\
\color{seedblue}Li Sun\\
\color{seedblue}Renfei Sun\\
\color{seedblue}Yan Sun\\
\color{seedblue}Zeyu Sun\\
\color{seedblue}Wenjing Tang\\
\color{seedblue}Yaxue Tang\\
\color{seedblue}Zirui Tao\\
\color{seedblue}Feng Wang\\
\color{seedblue}Furui Wang\\
\color{seedblue}Jinran Wang\\
\color{seedblue}Junkai Wang\\
\color{seedblue}Ke Wang\\
\color{seedblue}Kexin Wang\\
\color{seedblue}Qingyi Wang\\
\color{seedblue}Rui Wang\\
\color{seedblue}Sen Wang\\
\color{seedblue}Shuai Wang\\
\color{seedblue}Tingru Wang\\
\color{seedblue}Weichen Wang\\
\color{seedblue}Xin Wang\\
\color{seedblue}Yanhui Wang\\
\color{seedblue}Yue Wang\\
\color{seedblue}Yuping Wang\\
\color{seedblue}Yuxuan Wang\\
\color{seedblue}Ziyu Wang\\
\color{seedblue}Guoqiang Wei\\
\color{seedblue}Wanru Wei\\
\color{seedblue}Di Wu\\
\color{seedblue}Guohong Wu\\
\color{seedblue}Hanjie Wu\\
\color{seedblue}Jian Wu\\
\color{seedblue}Jie Wu\\
\color{seedblue}Ruolan Wu\\
\color{seedblue}Xinglong Wu\\
\color{seedblue}Yonghui Wu\\
\color{seedblue}Ruiqi Xia\\
\color{seedblue}Liang Xiang\\
\color{seedblue}Fei Xiao\\
\color{seedblue}XueFeng Xiao\\
\color{seedblue}Pan Xie\\
\color{seedblue}Shuangyi Xie\\
\color{seedblue}Shuang Xu\\
\color{seedblue}Jinlan Xue\\
\color{seedblue}Shen Yan\\
\color{seedblue}Bangbang Yang\\
\color{seedblue}Ceyuan Yang\\
\color{seedblue}Jiaqi Yang\\
\color{seedblue}Runkai Yang\\
\color{seedblue}Tao Yang\\
\color{seedblue}Yang Yang\\
\color{seedblue}Yihang Yang\\
\color{seedblue}ZhiXian Yang\\
\color{seedblue}Ziyan Yang\\
\color{seedblue}Songting Yao\\
\color{seedblue}Yifan Yao\\
\color{seedblue}Zilyu Ye\\
\color{seedblue}Bowen Yu\\
\color{seedblue}Jian Yu\\
\color{seedblue}Chujie Yuan\\
\color{seedblue}Linxiao Yuan\\
\color{seedblue}Sichun Zeng\\
\color{seedblue}Weihong Zeng\\
\color{seedblue}Xuejiao Zeng\\
\color{seedblue}Yan Zeng\\
\color{seedblue}Chuntao Zhang\\
\color{seedblue}Heng Zhang\\
\color{seedblue}Jingjie Zhang\\
\color{seedblue}Kuo Zhang\\
\color{seedblue}Liang Zhang\\
\color{seedblue}Liying Zhang\\
\color{seedblue}Manlin Zhang\\
\color{seedblue}Ting Zhang\\
\color{seedblue}Weida Zhang\\
\color{seedblue}Xiaohe Zhang\\
\color{seedblue}Xinyan Zhang\\
\color{seedblue}Yan Zhang\\
\color{seedblue}Yuan Zhang\\
\color{seedblue}Zixiang Zhang\\
\color{seedblue}Fengxuan Zhao\\
\color{seedblue}Huating Zhao\\
\color{seedblue}Yang Zhao\\
\color{seedblue}Hao Zheng\\
\color{seedblue}Jianbin Zheng\\
\color{seedblue}Xiaozheng Zheng\\
\color{seedblue}Yangyang Zheng\\
\color{seedblue}Yijie Zheng\\
\color{seedblue}Jiexin Zhou\\
\color{seedblue}Jiahui Zhu\\
\color{seedblue}Kuan Zhu\\
\color{seedblue}Shenhan Zhu\\
\color{seedblue}Wenjia Zhu\\
\color{seedblue}Benhui Zou\\
\color{seedblue}Feilong Zuo\\
\end{multicols} 

% \noindent
% \sffamily{\color{seedblue}  \large{Contributors}} \\
% \\
% % \color{black}Sheng Bi\\
% % \color{black}Hao Chen\\
% % \color{black}Haoshen Chen\\
% % \color{black}Haoxin Chen\\
% % \color{black}Xiaoya Chen\\
% % \color{black}Feng Cheng\\
% % \color{black}Xuyan Chi\\
% % \color{black}Xiaojing Dong\\
% % \color{black}Junliang Fan\\
% % \color{black}Jing Fang\\
% % \color{black}Liangke Gui\\
% % \color{black}Qiushan Guo\\
% % \color{black}Bibo He\\
% % \color{black}Ruoqing Hu\\
% % \color{black}Siqi Jiang\\
% % \color{black}Ashley Kim\\
% % \color{black}Gen Li\\
% % \color{black}Yiying Li\\
% % \color{black}Haibin Lin\\
% % \color{black}Feng Ling\\
% % \color{black}Gaohong Liu\\
% % \color{black}Zuxi Liu\\
% % \color{black}Zhibei Ma\\
% % \color{black}Yanghua Peng\\
% % \color{black}Lei Shi\\
% % \color{black}Zuquan Song\\
% % \color{black}Renfei Sun\\
% % \color{black}Qinlong Wang\\
% % \color{black}Xuanda Wang\\
% % \color{black}Xun Wang\\
% % \color{black}Ye Wang\\
% % \color{black}Meng Wei\\
% % \color{black}Yawei Wen\\
% % \color{black}Ruolan Wu\\
% % \color{black}Xiaohu Wu\\
% % \color{black}Yonghui Wu\\
% % \color{black}Xin Xia\\
% % \color{black}Tingshuai Yan\\
% % \color{black}Zhouqike Yang\\
% % \color{black}Ziyan Yang\\
% % \color{black}Linxiao Yuan\\
% % \color{black}Zhonghua Zhai\\
% % \color{black}Manlin Zhang\\
% % \color{black}Xinyan Zhang\\
% % \color{black}Xinyu Zhang\\
% % \color{black}Zixiang Zhang\\
% % \color{black}Qi Zhao\\
% % \color{black}Rui Zhu\\
% % \color{black}Wenjia Zhu
% \end{multicols} %

\end{CJK*}
\end{document}